\RequirePackage{fix-cm}
\documentclass[twocolumn]{svjour3}          
\smartqed  
\usepackage{graphicx}
\usepackage{subcaption}
\usepackage{amsmath}
\usepackage{algorithm}
\usepackage[noend]{algpseudocode}
\journalname{Preprint}
\begin{document}

\title{Convolutional Fully-Connected Capsule Network (CFC-CapsNet): A Novel and Fast Capsule Network}



\author{Pouya Shiri         \and
        Amirali Baniasadi 
}


\institute{P. Shiri \at
              Electrical and Computer Engineering Faculty
              University of Victoria \\
              \email{pouyashiri@uvic.ca}           
           \and
           A. Baniasadi \at
              Electrical and Computer Engineering Faculty
              University of Victoria \\
              \email{amiralib@uvic.ca}           
}

\date{Preprint}

\maketitle

\begin{abstract}
    A Capsule Network (CapsNet) is a relatively new classifier and one of the possible successors of Convolutional Neural Networks (CNNs). CapsNet maintains the spatial hierarchies between the features and outperforms CNNs at classifying images including overlapping categories. Even though CapsNet works well on small-scale datasets such as MNIST, it fails to achieve a similar level of performance on more complicated datasets and real applications. In addition, CapsNet is slow compared to CNNs when performing the same task and relies on a higher number of parameters. In this work, we introduce Convolutional Fully-Connected Capsule Network (CFC-CapsNet) to address the shortcomings of CapsNet by creating capsules using a different method. We introduce a new layer (CFC layer) as an alternative solution to creating capsules. CFC-CapsNet produces fewer, yet more powerful capsules resulting in higher network accuracy. Our experiments show that CFC-CapsNet achieves competitive accuracy, faster training and inference and uses less number of parameters on the CIFAR-10, SVHN and Fashion-MNIST datasets compared to conventional CapsNet.
\end{abstract}


\section{Introduction}
\label{intro}

Convolutional Neural Networks (CNNs) have achieved high performance in various applications including object detection and image classification \cite{Zhao2017}. Despite the advantages, CNNs have some limitations. These networks do not maintain the spatial hierarchies among the features of the image and have limited equivariance. These disadvantages are partly due to the loss of information occurring at  the max-pooling layers used in CNNs. Capsule Network (CapsNet) was introduced to address these drawbacks \cite{Sabour2017}. The structure of CapsNet and how it works is explained in detail in Section 2. Experiments show that CapsNet supports a higher level of equivariance and is also more effective than CNNs at detecting overlapping images \cite{Sabour2017}. It also does not use a max-pooling layer, as it builds on a different  structure. CapsNet maintains the spatial hierarchies between the features by creating a representation based on the part-to-whole relationship. The basic computational units in CapsNet are capsules: vectors of neurons. 

\par
CapsNet works well on the MNIST dataset \cite{LECUN}. However, the performance of CapsNet deteriorates as datasets become more complex, and it is still not useful for real applications. For example, CapsNet obtains a low accuracy of 18\% accuracy on CIFAR-100 dataset (a dataset more complex than MNIST)  while CNNs have achieved 75.72\% \cite{Mukhometzianov2018,Benenson2016}. In addition, the process of inferring capsules in subsequent layers is a computationally intensive method and this makes CapsNet relatively slow compared to CNNs \cite{Mobiny2018}.

\par
Sabour et al. introduced the first version of Capsule Networks \cite{Sabour2017}. In the first version of CapsNet, capsules are created out of low-level features extracted by two consecutive convolutional layers. The created capsules are multiplied by a matrix and the first layer of capsules is hence formed. These capsules are hierarchically processed  in consecutive capsule layers. In each layer, capsules are inferred from the previous layer using an iterative algorithm referred to as Dynamic Routing (DR), which is based on the agreement between the capsules of the previous layer. The capsules in the last layer (referred to as output capsules in this work) are fed to a simple decoder network consisting of Fully-Connected (FC) layers. This is done to reconstruct the input images and regularize the training process. 
\par
In this work we address the shortcomings of CapsNet by investigating and enhancing CapsNet's computational structure. The first layer of capsules also called Primary Capsules (PCs), is created from the output feature maps of the low-level feature extractor, reshaped to vectors. CapsNet then processes the created capsules in multiple layers. As there is a high number of capsules in the original architecture of CapsNet (1152 for the MNIST dataset), we performed some experiments to explore if the number of capsules could be reduced. We investigated how the number of PCs affects the network's performance by altering the number of feature maps of the second convolutional layer. We realized the number of PCs plays a significant role in the network performance. As a result, we came up with an alternative method for creating capsules. The experiments regarding the motivation, are explained in detail in the Experiments and Results section.

\par
 We introduce the Convolutional Fully-Connected (CFC) layer as an efficient mechanism to translate the extracted features to capsules. This layer, creates significantly fewer number of capsules, while leading to a more powerful representation. By integrating the CFC layer into CapsNet, we introduce CFC-CapsNet. CFC-CapsNet also employs a more robust decoder instead of the basic one used in CapsNet. This alternative, consists of deconvolution layers instead of FC layers. In addition, a different strategy is employed to feed the output capsules to the decoder.

CFC-CapsNet has several advantages over CapsNet. In summary, the main contributions of this network include:
\begin{itemize}
    \item \textbf{Improved network accuracy}: CFC-CapsNet creates the capsules out of the feature maps using a translation which is trained and it improves as the training makes progress. The capsules created using this method lead to a more powerful representation compared to conventional CapsNet. For instance, on the CIFAR-10 dataset the test accuracy is improved by 2.16\%.
    \item \textbf{Reduced number of parameters}: CFC-CapsNet creates a significantly fewer number of capsules. As we show later, the number of capsules is a key factor in determining the number of parameters in the network. Also, we employ an alternative decoder, which includes significantly fewer number of parameters as it uses deconvolution instead of FC layers. CFC-CapsNet reduces the number of parameters by 49\% on CIFAR-10 dataset. 
    \item \textbf{Speeding up the network}: Reducing the number of capsules, speeds up the training and the inference of the network. Using fewer capsules, it takes less time for inferring capsules in the subsequent layers  to find an agreement and produce the capsules for the next layer. Compared to CapsNet, CFC-CapsNet is 4x and 4.5x faster on the CIFAR-10 dataset for the training and inference respectively.
\end{itemize}

The rest of the paper is organized as follows. We review the background in Section 2. Section 3 discusses the related works. Section 4 explains our proposed architecture. Section 5 reports the experimental results. We offer concluding remarks in in Section 6.

\section{Background}

\subsection{Capsule Network}
\begin{figure*}
    \centering
    \includegraphics[width=\textwidth,height=\textheight,keepaspectratio]{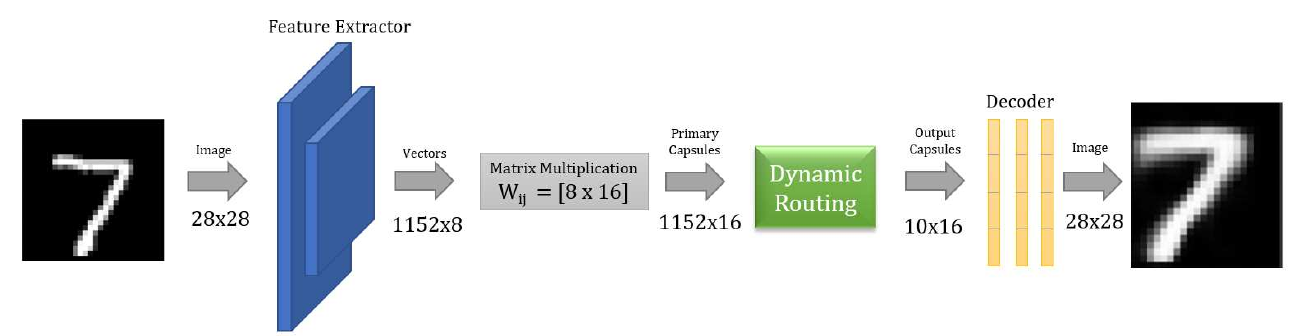} 
    \caption{The architecture of CapsNet. After extracting low-level features using two convolutional layers, the output is reshaped to vectors. These vectors are multiplied by a matrix to create the primary capsules which constitute the basic unit of CapsNet. The output capsules are inferred using the dynamic routing algorithm, and finally the decoder reconstructs the input image.}
    \label{fig:back_1}
\end{figure*}
\begin{figure*}
    \centering
    \includegraphics[width=\textwidth,height=\textheight,keepaspectratio]{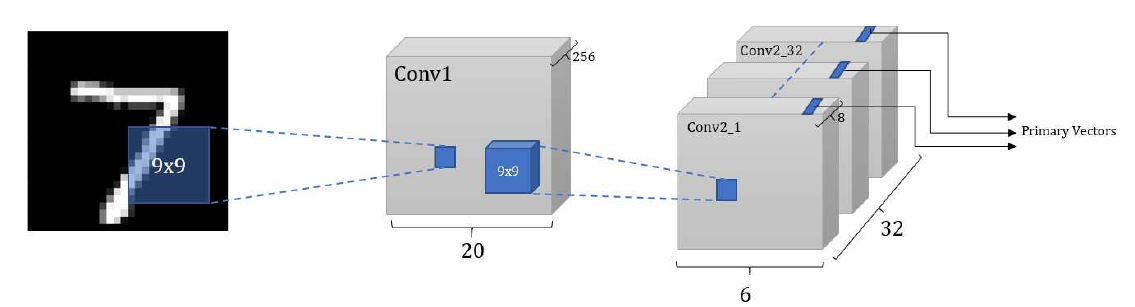} 
    \caption{The low-level feature extractor in CapsNet. There are two convolutional layers. The output is reshaped to vectors. }
    \label{fig:caps_feature_ext}
\end{figure*}

The basic computational units in CapsNet are capsules. A capsule is a group of neurons structured as a vector. Figure \ref{fig:back_1} shows the architecture of CapsNet when applied to the MNIST dataset \cite{LECUN}.  As the figure shows, CapsNet starts with a feature extractor. This is illustrated in Figure \ref{fig:caps_feature_ext}. This is a low-level feature extractor that takes as input the image, and produces vectors. Each spatial position of the output feature map in the second layer, is taken as a vector.

 The output of the feature extractor (the created vectors, $u_i$) are then multiplied by weight matrices ($W_{ij}$ that encodes the spatial relationship between the vectors:
 \begin{equation}
     \hat{u}_{j|i} = W_{ij}u_i 
 \end{equation}

 The result is the first layer of capsules, which is referred to as the Primary Capsules (PCs, $\hat{u}_{j|i}$). 
\par
The output capsules are inferred from the primary capsules using the Dynamic Routing (DR) algorithm.  All primary capsules contribute to all output capsules. The DR algorithm is in charge of finding the corresponding coefficient for each of the primary capsules. In other words, it finds a suitable routing between the input capsules and the capsules in the next layer. These coefficients are not trainable parameters, and the DR algorithm calculates them at every iteration during the training process based on the agreement between the primary capsules. 
\par
Algorithm 1 shows the pseudo-code for the DR method. The inputs of this algorithm are the primary capsules ($\hat{u}_{j|i}$) in a lower level $l$ and the number of routing iterations $r$. The algorithm determines the values for the coefficients $c_{ij}$ and creates the output capsules $v_j$. $b_{ij}$s are temporary values which are later used to determine the coefficients. The first step (line 4), is to calculate the $softmax$ function for $b$ values to enforce the probablistic nature of the coefficients based on the following equation:

\begin{equation}
    c_{ij} = \frac{exp(b_{ij})}{\sum_k exp(b_{ik})}
\end{equation}

Afterwards, the output is created based on the calculated coefficients (line 5). The non-linear $squash$ function is used to shrink short vectors to zero length, and long vectors to a length of slightly below 1 (line 6). This function is based on the following equation:

\begin{equation}
    v_j = \frac{{||s_j||}^2}{1+{||s_j||}^2}\frac{s_j}{||s_j||}
\end{equation}

Finally (line 7), the temporary variables $b$ are updated based on the inner product between the current output and each capsule. The inner product is used a good measure of level of agreement between the input capsules and the output.

\begin{algorithm}

\caption{The Dynamic Routing Algorithm \cite{Sabour2017}}
\begin{algorithmic}[1]

\Procedure{Routing}{$\hat{u}_{j|i},r,l$}       
    \State for all capsule $i$ in layer $l$ and capsule $j$ in layer $(l+1)$: $b_{ij} \leftarrow 0$.

    \For{\texttt{$r$ iterations}}
        \State for all capsule $i$ in layer $l$: $c_i \leftarrow softmax(b_i)$
        \State for all capsule $j$ in layer $(l+1)$: $s_j \leftarrow \sum_{i} c_{ij}{\hat{u}}_{j|i}$
        \State for all capsule $j$ in layer $(l+1)$: $v_j \leftarrow squash(s_j)$
        \State for all capsule $i$ in layer $l$ and capsule $j$ in layer $(l+1)$: $b_{ij} \leftarrow b_{ij}$ + ${\hat{u}}_{j|i}.v_j$
      \EndFor

    \State \textbf{return} $v_j$
\EndProcedure

\end{algorithmic}
\end{algorithm}

\par
\begin{figure*}
    \centering
    \includegraphics[width=\textwidth,height=\textheight,keepaspectratio]{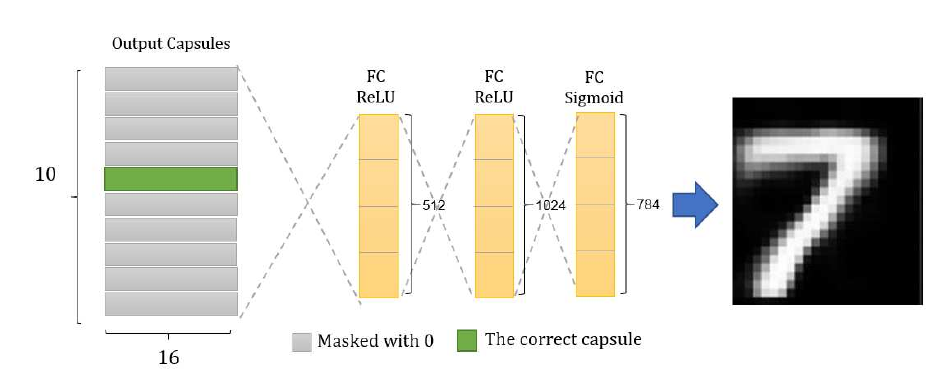} 
    \caption{CapsNet Decoder network. The image is reconstructed using subsequent FC layers. Note that only the correct capsule (the capsule with the correct prediction) is kept and the rest of the capsules are masked with zeros.}
    \label{fig:caps_dec}
\end{figure*}

 The total loss function of CapsNet includes the Margin Loss (explained later in this section) and the Reconstruction Loss. CapsNet includes a decoder network to reconstruct the input images to encourage the output capsules to encode the input image. The decoder is used to regularize the training process. As Figure \ref{fig:caps_dec} shows, this network receives the output capsules as input and uses subsequent Fully-Connected (FC) layers. The sum of absolute differences between the reconstructed and the input image, creates a reconstruction loss term. This term would be higher if the reconstructed images are less similar to the input images. This is a common regularization technique to avoid overfitting. The decoder can further be used after training the network to verify how each dimension of the output capsule encodes the input image. This is done by adding noise to the output capsules of a trained network, and checking how the reconstructed images change \cite{Sabour2017}.

\par

There are as many capsules as the number of categories in the classification task. The images are classified using the output capsules: the length (L2-Norm) of each capsule shows the probability of the image belonging to the category corresponding to that capsule. In addition capsules hold other important information. Ideally, each dimension of each capsule corresponds to an instantiation parameter e.g. deformation, pose (size, orientation and position), texture, etc. 

\par
The loss function in CapsNet (called margin loss) consists of penalties considered for the output mispredictions of the network. The predictions of all output capsules impact the margin loss. The loss function consists of the following term for each output capsule: 
\begin{equation}  \resizebox{0.91\hsize}{!}{$L_k = T_k \max(0,m^+-||V_k||)^2 + \lambda(1-T_k)\max(0,||V_k||-m^-)^2$}
\end{equation}

where $T_k$ is 0 if the prediction of the k-th capsule is incorrect and 1 otherwise, $||V_k||$ is the magnitude of the k-th output capsule, $\lambda$ is the down-weighting coefficient for the incorrect predictions, and $m^+$ and $m^-$ are considered to dismiss predictions with a very high (and very low) probability from participating in the loss term.

\subsection{Class-Independent Decoder}
Rajasegaran et al. propose an alternative decoder for CapsNet referred to as the class-independent decoder \cite{Rajasegaran}. This decoder uses deconvolution layers instead of the FC layers used in CapsNet. Deconvolutional layers are more powerful than FC layers in capturing spatial relationships \cite{Rajasegaran}. The other advantage of this decoder is the reduced number of parameters. Similar to convolutional layers, deconvolutional layers also take advantage of the weight-sharing property. As a result, these layers are lighter and include fewer number of parameters. 

\par

\begin{figure*}
    \centering
    \includegraphics[keepaspectratio,scale=0.8]{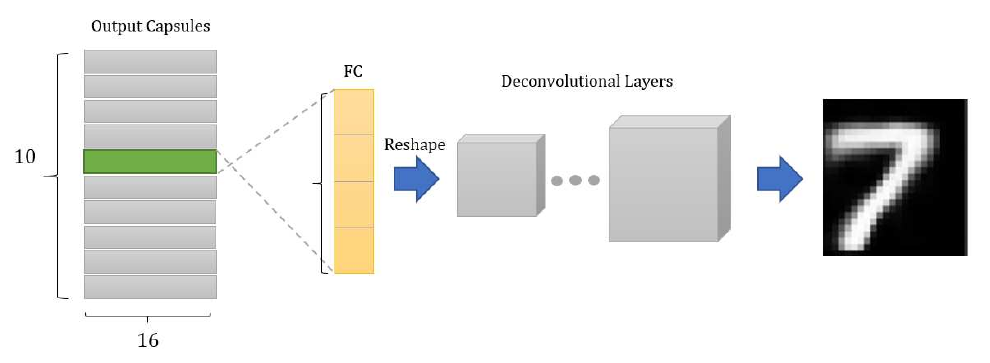} 
    \caption{Class-Independent Decoder for CapsNet. }
    \label{fig:caps_newdec}
\end{figure*}

\par
 As Figure \ref{fig:caps_dec} shows, in the original decoder all capsules are masked out with zeros except for the capsule with the correct prediction, which is determined based on the ground-truth labels during the training. During testing, the capsule with the highest magnitude corresponds to the correct prediction. Therefore, this capsule is kept and the rest of capsules are masked with zeros. All of the capsules (both the masked ones and the capsule with the correct prediction) are fed to the decoder network. The way this method works only depends on the correct category, because the input of the decoder is constructed using the information from all categories while the incorrect categories are masked. Making the decoder independent of the correct category makes the decoder more robust. 
 
 \par
 Figure \ref{fig:caps_newdec} shows the architecture of the class-independent decoder proposed by \cite{Rajasegaran}. The correct capsule is fed to an FC layer followed up by several deconvolutional layers which expand the representation progressively. This decoder is independent of the image label (category) as it only  considers a single capsule as the input of the reconstruction network: the capsule with the correct prediction. Hence we refer to it as a class-independent decoder. In this decoder, different dimensions of the capsule are treated the same for different categories.

\section{Related Works}

There are several studies that have improved CapsNet. Some studies focus on modifying the feature extractor and the decoder network. Rosario et al. propose a resource-efficient version of CapsNet called Multi-Lane Capsule Network (MLCN) \cite{DoRosario2019}. This network is more accurate than CapsNet and is faster during training and inference. It also considers a different "lane" to generate each dimension of the output capsule. Lanes have various depths and widths: the depth corresponds to the number of convolutional layers and the width corresponds to the number of filters used per convolutional layer. MLCN also incorporates a novel dropout technique referred to as the  lane dropout. This method randomly drops 10\% of lanes while training to make the representation independent of specific lanes.

\par
Pouya et al. \cite{shiri2020} proposed Quick-CapsNet (QCN) as a fast variant of CapsNet. They change the architecture of the feature extractor to integrate a Fully-Connected layer instead of the second convolutional layer. This resulted in a significantly faster training and inference. The default decoder of CapsNet is also replaced with a more powerful one based on deconvolution. This network reduces the number of PCs significantly and achieves a competitive accuracy. 

\par
Another variant of CapsNet achieving high accuracy, is Multi-Scale Capsule Network (MS-CapsNet) introduced by Xian et al. \cite{Xiang2018}. The feature extractor in this network consists of several scales. The extracted features in each scale are converted to vectors. There are different matrices multiplied by the output vectors in each scale to create predictions. The resulting predictions are concatenated to create the primary capsules. MS-CapsNet also introduces a capsule dropout method to regularize the training process. In this method, some capsules are randomly dropped while training to make the representation independent of specific capsules. 

\par

In \cite{Rajasegaran}, Rajasegaran et al. propose DeepCaps. In this work the feature extractor of DeepCaps is deeper than that of CapsNet and includes several skip-connected convolutional layers. This network also proposes a novel DR algorithm that is based on 3D convolution and a more powerful class-independent decoder, which uses deconvolution layers instead of FC layers used in the original decoder and therefore creates more powerful reconstructions.

\par
Another deep variant of CapsNet is Deep-Tensor CapsNet \cite{Sun2020}. This network introduces tensor capsules and introduces a new DR algorithm based on tensor capsules. Also, to solve the problem of overfitting, this network regularizes the training process by proposing a dropout method for vectors and tensors. There is a multi-scale decoder in this network that provides more clear reconstructed images compared to the original decoder.

\par
Deliege et al. \cite{Deli2018} propose HitNet as a novel CapsNet that integrates a Hit-or-Miss layer in it. This layer consists of capsules used to trigger a central capsule. This network uses a modified loss function to handle the newly introduced layer. HitNet also includes a reconstruction network that can create samples of images. This could be used for augmenting the training set and regularize the training process. The other novelty in HitNet, is introducing the concept of ghost capsules, which detect the data wrongly labeled in the training set. 

\par
Fuchs et al. \cite{Fuchs2020} introduce a new variant of CapsNet that is capable of classifying complex datasets. This network uses a Wasserstein objective. This objective is used to select capsules for the routing dynamically. The new routing process results in a little overhead. The introduced network obtains a lower classification error and also reduces the number of parameters.

\par
He et al. \cite{He2019} propose two new architectures based on CapsNet called Cv-CapsNet and Cv-CapsNet++. These networks use a complex-valued decoder to extract low-level features in multiple scales. The extracted features are then used to form complex-valued primary capsules. The dynamic routing algorithm of CapsNet is modified to work in the complex domain. 

\par
Yang et al. \cite{Yang2020} introduce RS-CapsNet, which uses Res2Net blocks for multi-scale feature extraction. This network also consists of Squeeze-and-Excitation (SE) blocks to scale the features based on importance. A linear combination of capsules is used to reduce the number of capsules and create capsules with a higher representation ability. RS-CapsNet also introduces intermediate capsules, mostly responsible for representing the detected object. These capsules contribute to the primary capsules in the task of classification.

\par
Huang et al. propose DA-CapsNet \cite{Huang2020}. This network uses the attention mechanism both after the low-level feature extractor and the primary capsules. DA-CapsNet performs a better reconstruction and obtains state-of-the-art accuracy on small-scale datasets e.g. SVHN and CIFAR-10.

\par
There are several other studies that focus on the DR algorithm and offer alternatives to this method. Hinton et al. introduce matrix capsules in \cite{Hinton2018b}. Matrix capsules keep the pose information in a 4x4 matrix, keeping the viewpoint information of the object. The alternative routing mechansim used for matrix capsules is referred to as the Expectation-Maximization (EM) routing. In this method, the input capsules are considered as datapoints and are fitted to a mixture of Gaussian distributions. Each distribution corresponds to one of the output capsules. This network has significantly fewer number of parameters and performs better than CapsNet in terms of the classification accuracy.

\par
Another network with an alternative routing algorithm is Group Feedback Capsule Network (GF-CapsNet) \cite{Ding2020}. This work tries to reduce the high computation complexity of the original dynamic routing method. This is done by dividing capsules into groups sharing the same transformation weights. This reduces the number of parameters required for the routing. In addition, authors mention that the the conventional routing mechanisms include some distribution assumptions that might not hold true for real-world data. To deal with this issue, GF-CapsNet includes a distance network for predicting capsules directly in a supervised manner.

\par
In Straight-Through Attentive Routing Capsule Network (STAR-CAPS), Ahmed et al. \cite{Ahmed2019} utilizes the attention modules to estimate the coefficients of the routing algorithm. These modules are augmented by binary routers, and the presented alternative routing method has no recurrence. The straight-through estimators make a binary decision on the connection of each input capsules to the each output capsule, and this leads to a more stable and faster performance.

\par
Some works on CapsNet focus on improving the network for a specific applications or datasets. Zhang et al. propose an enhanced capsule network \cite{Zhang2020} for medical image classification. This network includes a feature decomposition module along with a multi-scale feature extractor. These modules are used to extract features more efficiently, reducing the number of calculations and speeding up the convergence. The proposed network was applied to the PatchCamelyon (PCam) dataset and obtained good performance.

\par
Mobiny et al. propose the Detail-Oriented Capsule Network (DECAPS) \cite{Mobiny2020} which is specialized for medical imaging datasets. This network includes some novel methods to increase the classification accuracy. DECAPS includes an alternative method for routing capsules called the Inverted Dynamic Routing (IDR). This method groups capsules in the lower level before routing to the higher-level capsules. This novel routing intensifies the small but informative details within the data. The other innovation of this work, is the Peekaboo training procedure which includes an attention scheme to focus on fine grain information. In addition, the robustness of the network is improved by averaging the attended and the original image predictions. DECAPS is tested on the CheXpert and RSNA Pneumonia datasets and it obtains state-of-the-art accuracy.

\par
Mobiny et al. propse a fast variant of CapsNet \cite{Mobiny2018a} specifically designed for lung cancer detection. This network proposes a consistent dynamic routing method that results in speeding up the CapsNet. In addition, since the original reconstruction network works poorly on lung data, a convolutional decoder is used instead, resulting in higher reconstruction and classification accuracy.

\par
Table \ref{table:trel_sum} summarizes the related works we presented in this paper. Most of the works modify the feature extractor and the reconstruction network (the decoder). Some works present a modified method for routing low-level capsules to high-level ones and some works introduce new mechanisms and methods to create capsules with a higher representation capability e.g. using the attention mechanism. Three related works focus on improving CapsNet for specific domain (medical imaging) while others focus on making improvement on the conventional small-scale datasets. 

\par
This work is different from the works mentioned above, as it focuses on how the primary capsules are generated. Most of the proposed networks based on CapsNet, simply reshape the extracted features to capsules and this leads to a high number of capsules. In this work we build on what we presented earlier \cite{Shiri2021}, where we introduced the Convolutional Fully-Connected (CFC) layer as a feature summarization method that leads to generating only a few capsules. We further investigate this layer by exploring the different choice of parameters, and propose ideas for a better integration of this layer into CapsNet. The primary focus of this work is to provide a lighter and faster variant of CapsNet by summarizing the extracted features using a special method which is based on capsules. There are several general methods for summarizing the network. However, most of these methods are offline. In other words, they modify the network after the training is complete. For example, pruning methods attempt to simplify the network after training is complete by removing neurons based on different criteria to provide a lighter network for the inference \cite{Molchanov2017}. Polino et al. \cite{Polino2018} propose two model compression methods based on weight quantization and distilling larger networks into small student networks. 

\begin{table*}[htp]
\caption{A summary of related works focused on CapsNet} 
\centering 
\resizebox{\linewidth}{!}{
\begin{tabular}[width=\linewidth]{|l|l| c|}
 \hline
\textbf{Model} & \textbf{Innovation} & \textbf{Target Datasets} \\ [0.5ex] 
\hline 
\hline
MLCN \cite{DoRosario2019} & Multiple lanes with different depth and width, Lane dropout & Small-Scale \\ [1ex] 
\hline
QCN \cite{shiri2020} & Modifying the feature extractor and the decoder, reducing PCs & Small-Scale \\ [1ex]
\hline
MS-CapsNet \cite{Xiang2018} & Modifying the feature extractor, Capsule Dropout & Small-Scale \\[1ex]
\hline
DeepCaps \cite{Rajasegaran2019} & Deep feature extraction, 3D Dynamic Routing, improved decoder & Small-Scale \\[1ex]
\hline
DeepTensor \cite{Sun2020} & Introducing tensor capsules, modified dynamic routing, vector dropout & Small-Scale \\[1ex]
\hline
HitNet \cite{Deli2018} & Introducing Hit-and-Miss layer and ghost capsules & Small-Scale \\[1ex]
\hline
WR-CapsNet \cite{Fuchs2020} & Using Wasserstein objective, modifying dynamic routing & Small-Scale \\[1ex]
\hline
Cv-CapsNet \cite{He2019} & Introducing complex-valued capsules and dynamic routing & Small-Scale \\[1ex]
\hline
RS-CapsNet \cite{Yang2020} & Integrating Res2Net and Squeeze-and-Excitation blocks, and intermediate capsules & Small-Scale \\[1ex]
\hline
DA-CapsNet \cite{Huang2020} & Integrating Attention mechanism in the feature extractor and primary capsules & Small-Scale \\[1ex]
\hline
EM-CapsNet \cite{Hinton2018b} & Introducing Matrix capsules, Applying the EM method for routing capsules & Small-Scale \\[1ex]
\hline
GF-CapsNet \cite{Ding2020} & Grouping capsules, modified routing, reducing the required parameters for routing & Small-Scale \\[1ex]
\hline
STAR-CAPS \cite{Ahmed2019} & Estimating routing coefficients, modified routing & Small-Scale\\[1ex]
\hline
Enhanced CapsNet \cite{Zhang2020} & Using feature decomposition module and a multi-scale feature extractor & PCam \\[1ex]
\hline
DECAPS \cite{Mobiny2020} & Inverted dynamic routing (IDR), grouping capsules, Peekaboo training & CheXpert, RSNA \\[1ex]
\hline
CapsNet for Lung Cancer \cite{Mobiny2018a} & modified dynamic routing and the decoder & Lung Cancer Dataset \\[1ex]

\hline 
\end{tabular}}
\label{table:trel_sum} 
\end{table*}

\section{Convolutional Fully-Connected CapsNet}

In this section we explain our proposed network. First we explain the Convolutional Fully-Connected (CFC) layer and how it summarizes the feature map into significantly fewer number of capsules. Next, we explain the decoder we have used in this work.

\par
\subsection{Convolutional Fully-Connected Layer}
\begin{figure}[htp]
    \centering
    \includegraphics[keepaspectratio,scale=0.5]{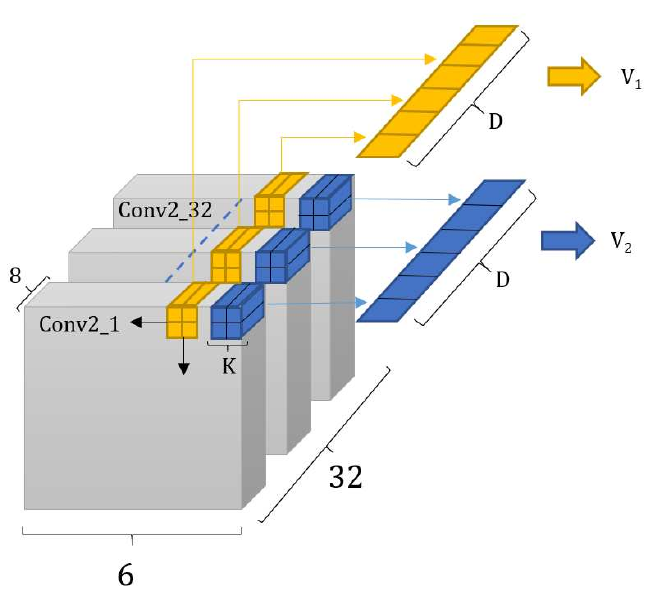} 
    \caption{CFC Layer. In order to provide a summary of vectors, all the vectors corresponding to spatially correlated neurons are grouped together and fed to different FC layers. Kernel size (K) and output dimensionality (D) are shown in the Figure. }
    \label{fig:cfc_layer}
\end{figure}
As mentioned in the Background section, the output of the low-level feature extractor is reshaped a high number of vectors. Here we propose a layer referred to as the Convolutional Fully-Connected (CFC) layer to translate the extracted features to vectors. As Figure \ref{fig:cfc_layer} shows, vectors corresponding to spatially correlated regions of the output activation are grouped together. The result is fed to different FC layers, each of which produces a single vector. 
\par
The CFC layer has two parameters: kernel size (K) and the output dimensionality (D). The kernel size determines the size of each vector group to be summarized into a single vector. The higher choice of K, results in fewer generated vectors. The output dimensionality determines the number of elements in each generated output vector. The number of output elements in FC layers determines the output dimensionality.
\par
The CFC layer is essentially the same as a convolutional layer, except for the fact that it does not implement the weight-sharing property. In convolutional layers, there is a single set of weights (the kernel) convolved with the input, whereas in the CFC layer, there are different kernels for each spatially correlated region of the input. 
\par

 Now we formulate the CFC layer using a mathematical notation. As mentioned before, there is a low-level feature extractor. The output activation of this extractor is used to create the capsules. The activation is split to different chunks ($C_m$) based on the following equation:
\begin{equation}
   C_m = (F_{a b c})_{\substack{a \in [w, w+K] \\ b \in [h, h+K] \\  c \in [1,N]}}
\end{equation}
where $m \in [1,(W-K+1)^2]$ and F is the output of the feature extractor, $F_{abc}$ shows the neuron at the $c$-th feature map of F at the spatial position $x=a$ and $y=b$, $N$ shows the number of feature maps, $K$ is the parameter of the CFC layer and $h$ and $w$ are obtained using the following equations:
\begin{equation}
    h = \frac{m}{W-K+1}
\end{equation}

\begin{equation}
    w = m \% (W-K+1)
\end{equation}
where the percentage denotes the remainder (mod) operation.

\par
Each chunk is then flattened. We refer to the flattened chunks as $C_Fm$. Each flattened chunk is then fed to a fully-connected layer with weights $W$ and the capsules are created based on the following equation:
\begin{equation}
V_m = C_Fm \cdot W_m
\end{equation}
where $V_m$ denotes each capsule including D values.

\subsection{Class-Independent Decoder}
We modify the class-independent decoder explained in the Background section by making it deeper and use it as the decoder of CFC-CapsNet. The deconvolutional layers used in this decoder are listed in Table \ref{table:tbl-dec-list}. It must be noted that for F-MNIST the output vector is projected to a 8x6x6 feature map. For SVHN and CIFAR-10, the output vector is projected to a 8x10x10 feature map to keep the decoder's structure consistent.

\begin{table}[htp]
\caption{Deconvolutional layers of the decoder} 
\centering 
\resizebox{150pt}{!}{\begin{tabular}[width=\linewidth]{|c|c| c|} 
\hline\hline 
\textbf{Layer} & \textbf{Input Dims} & \textbf{Kernel} \\ [0.5ex] 
\hline
1 & 8x6x6 & 128x3x3 \\
\hline
2 & 128x8x8 & 64x5x5 \\
\hline
3 & 64x12x12 & 32x5x5 \\
\hline
4 & 32x16x16 & 16x5x5 \\
\hline
5 & 16x20x20 & 16x5x5 \\
\hline
6 & 16x24x24 & 16x3x3 \\
\hline
7 & 16x26x26 & 1x3x3 \\
\hline

\end{tabular}}
\label{table:tbl-dec-list} 
\end{table}

\subsection{Capsule Dropout}

Following the capsule dropout method used in MS-CapsNet \cite{Xiang2018}, CFC-CapsNet includes a capsule dropout method for improving the generalization ability of the network by regularizing the training process. To this end, we randomly drop some of the primary capsules and follow the DR algorithm using the remaining capsules. It is noteworthy that we do not drop the elements inside vectors, as it might change the direction of the vector and affect the representation. Instead, we drop the whole capsule. We have a single experiment for each capsule, whether to drop it or to keep it. Hence, we can use the Bernoulli distribution: the disregarding capsules is considered as a success, and a failure otherwise. In our dropout layer, we implement the inverted dropout method. In other words, if $p$ is the drop probability, the primary capsules are scaled by $\frac{1}{1-p}$ during the training. All the neurons corresponding to these capsules are divided by $1-p$. This layer does not make any change to the capsules during inference. 

\subsection{Hard Training}
We use the hard training method used in the code provided by the authors of DeepCaps \cite{Rajasegaran}. In this method, first the network is trained using the default values for $m_+$ and $m_-$ in the loss function. After the network becomes stable, it is trained using the new values for $m_+$ and $m_-$. These new values are tighter than the previous ones: the new $m_+$ is higher than the previous one, and the new $m_-$ is lower than the previous one. In other words, the threshold for considering the predictions in the loss function is tightened and the training process is repeated. We chose the new values for $m_+$ and $m_-$ equal to the reference method ($m_+ = 0.95$ and $m_- = 0.05$). 

\section{Experiments and Results}
In this section, we explain our experiments and their results in details. The code for CFC-CapsNet is available on Github \footnote{https://github.com/pouyashiri/CFC-CapsNet.git}.
\subsection{Experiment Configurations}
\subsubsection{Datasets}
CapsNet and its variants are usually tested against small-scale datasets. Research studies do not use large image sizes and high number of categories for CapsNet, as the network becomes significantly bigger and slower as the number of categories grows. We perform experiments using Fashion-MNIST (F-MNIST) \cite{Xiao2017}, SVHN \cite{Netzer2011}, CIFAR-10 \cite{Krizhevsky2009}. To verify robustness against affine transformations, we test CFC-CapsNet on the AffNIST dataset. 
\par

\begin{table}[htp]
\caption{Datasets used to test CFC-CapsNet } 
\centering 
\resizebox{\linewidth}{!}{\begin{tabular}[width=\linewidth]{|c|c| c| c| c|c|} 
\hline \hline
\textbf{Name} & \textbf{Image Size} & \textbf{\#Channels} & \textbf{Training samples} & \textbf{Test Samples}  \\ [1ex]

\hline
MNIST & 28x28 & 1 & 50,000 & 10,000  \\ [1ex]
\hline
F-MNIST & 28x28 & 1 & 50,000 & 10,000 \\ [1ex]
\hline
SVHN & 32x32 & 3 & 73,257 & 26,032 \\ [1ex]
\hline
CIFAR-10 & 32x32 & 3 & 50,000 & 10,000  \\ [1ex]

\hline 
\end{tabular}}
\label{table:table_dsets} 
\end{table}

The datasets we used for this research and their properties are summarized in Table \ref{table:table_dsets}. Fashion-MNIST is a dataset following the same data structure as that of the MNIST digit recognition dataset \cite{LECUN}. However, instead of digits it contains images of different types of clothing, and hence it is considered a more complex and challenging dataset. It has 28x28 grey-scale images of 10 categories and the training and testing sets include 50,000 and 10,000 images respectively.

\par
SVHN and Cifar-10 dataset share the same data structure, but different total number of images. They both include 32x32 RGB images of 10 categories. There are 73,257 and 26,032 images in the training and testing sets of the SVHN dataset, while there are 50,000 and 10,000 images in the Cifar-10 dataset. The SVHN dataset contains cropped digits from the images of the house numbers. The Cifar-10 includes images of 10 different categories i.e. dogs, deer, cats, frogs, birds, cars, horses, ships, trucks and airplanes. The background varies a lot in the Cifar-10 dataset and it makes this dataset the most complex and challenging among the used datasets.

\subsubsection{Experiment Settings}
We implement CFC-CapsNet on top of the PyTorch implementation of CapsNet \footnote{https://github.com/gram-ai/capsule-networks}. Some experiments are performed using a Tesla T4 GPU with 16GB VRAM whereas others use a 2080Ti GPU with 11GB VRAM. We perform hard training in all experiments unless stated otherwise: the network is trained for 100 epochs (as there is no further significant change afterwards), and then trained for another 100 epochs using the new values for $m_+$ and $m_-$. All the experiments are repeated 5 times, and due to insignificant variance among the results, we report the average values. We use the Adam optimizer with the default learning rate ($LR=0.001$) and an exponential decay of $\gamma=0.96$ for the learning rate. The batch size is set to 128.

\subsection{Results}
In this section, first we explain the experiments motivating this work. We report the impacts of varying the number of PCs on network performance. Later, we provide a comparison between CFC-CapsNet and CapsNet under different configurations and using different metrics. We report network test accuracy, network training and inference times and the number of parameters. Furthermore, we explore the effect of the CFC layer parameters on the network (K and D). Then we find the best values of the parameters of the CFC layer based on the experiments and compare CFC-CapsNet and CapsNet using the selected values.

\subsubsection{Number of Primary Capsules}
These experiments are performed using the 2080Ti GPU without hard training. To investigate the effect of the number of PCs on different aspects of the network, we changed the number of kernels in the second convolutional layer.  We choose this number as $N_k \in \{32, 64, 128, 192, 256\}$. As each output vector is 8-dimensional and the size of the feature map in this case is 8x8, the number of capsules would be $N_{PC} \in \{256, 512, 1024, 1536, 2048\}$. We perform these experiments on CIFAR-10 dataset.

\par
\begin{figure}[htp]
    \centering
    \includegraphics[keepaspectratio, scale=0.5]{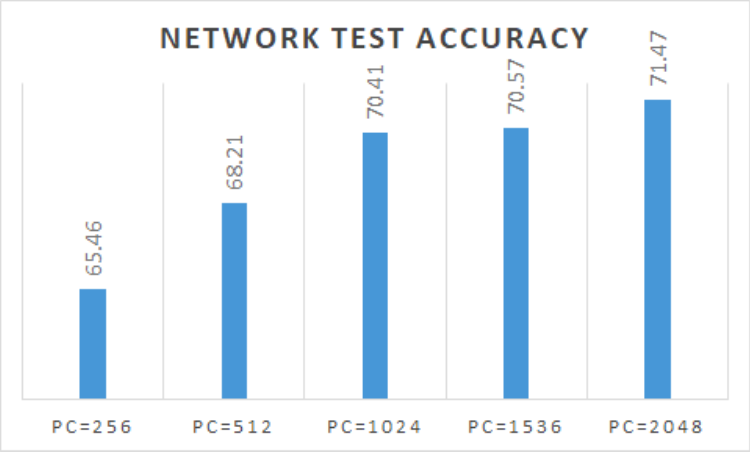}
    \caption{Network accuracy for different numbers of capsules}
    \label{fig:motiv_acc}
\end{figure}

\begin{figure}[htp]
    \centering
    \includegraphics[keepaspectratio, scale=0.5]{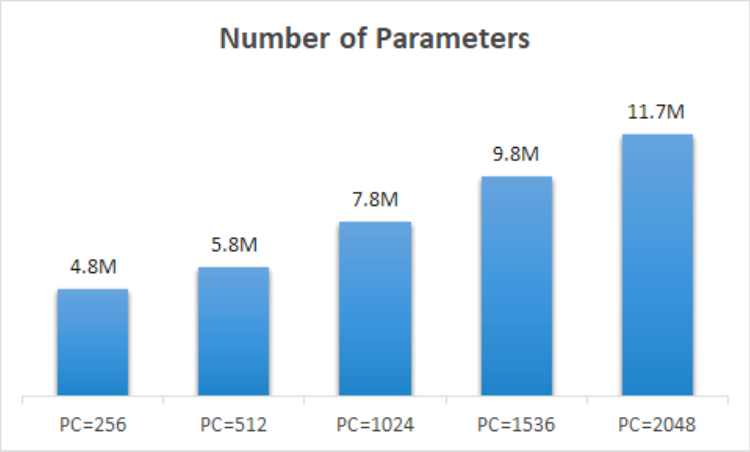}
    \caption{Network number of parameters for different numbers of capsules}
    \label{fig:motiv_params}
\end{figure}
As Figure \ref{fig:motiv_params} shows, the number of parameters changes significantly from 11.7M for 2048 PCs down to 4.8M for 256 PCs. This the significant effect of the number of capsules available in the network on the total number of parameters. 

\begin{figure}[htp]
    \centering
    \includegraphics[keepaspectratio,scale=0.5]{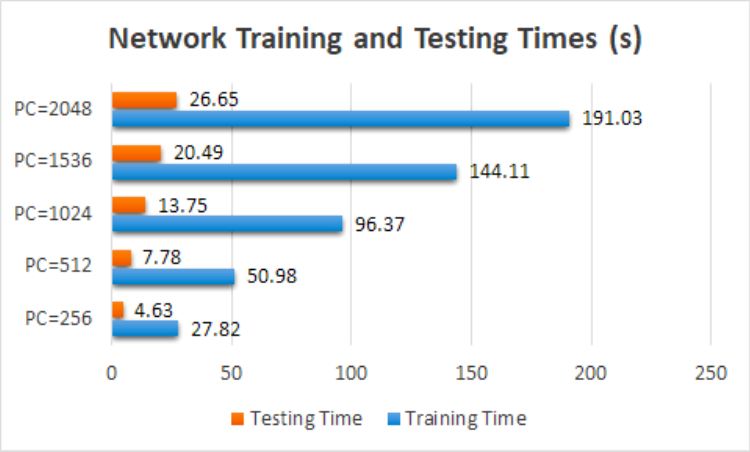}
    \caption{Network training and testing time for different numbers of capsules}
    \label{fig:motiv_speed}
\end{figure}
Finally, Figure \ref{fig:motiv_speed} shows how the network training and testing times change by varying the number of primary capsules. As this figure shows, the network is six times faster using 256 PCs compared to the case of 2048 capsules. 

\par
Considering the results for the different metrics, it is clear that the number of capsules plays a significant role in CapsNet. While reducing the number of capsules makes the network faster and lighter (fewer number of parameters), it drops the network accuracy and therefore is costly. We aim at coming up with an approach to reduce the number of capsules while maintaining the test accuracy. In CFC-CapsNet, we maintain the default number of kernels for the second convolutional layer in order to have the same amount of information. In the meantime, we use our solution to reduce the output size of the feature extractor to get fewer capsules.


\subsubsection{Parameter Exploration}
In this section we investigate using different values for the parameters of the CFC layer: D and K. We choose $D \in {8, 16, 32}$ and $K \in {1, 2, 3}$. We test the different parameters on the FMNIST and CIFAR-10 datasets and report network accuracy, the number of parameters and the inference time. We skip the SVHN dataset due to its similarity in structure and complexity to the CIFAR-10 dataset. For the experiments of this section, we do not perform hard training. This is because we are performing a comparison and not interested in maximizing accuracy. 

\par

\begin{figure*}
\centering
\begin{subfigure}{.5\linewidth}
  \centering
  \includegraphics[width=\linewidth]{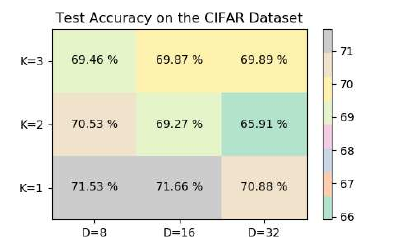}
  \caption{CIFAR-10 Accuracy}
  \label{fig:sub1}
\end{subfigure}%
\begin{subfigure}{.5\linewidth}
  \centering
  \includegraphics[width=\linewidth]{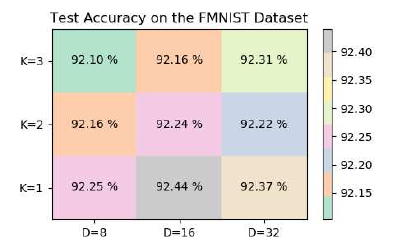}
  \caption{FMNIST Accuracy}
  \label{fig:sub2}
\end{subfigure}
\caption{CFC-CapsNet accuracy for different choices of K and D}
\label{fig:acc_param}
\end{figure*}
As Figure \ref{fig:acc_param} shows, for the FMNIST dataset there is not much difference in accuracy for different choices of K and D. For the CIFAR-10 dataset, $K=1$ is the best choice. For this choice, both $D=8$ and $D=16$  achieve competitive results.

\par

\begin{figure*}
\centering
\begin{subfigure}{.5\linewidth}
  \centering
  \includegraphics[width=\linewidth]{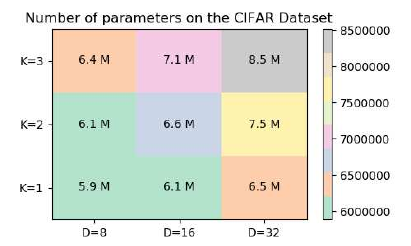}
  \caption{CIFAR-10 number of parameters}
  \label{fig:sub1}
\end{subfigure}%
\begin{subfigure}{.5\linewidth}
  \centering
  \includegraphics[width=\linewidth]{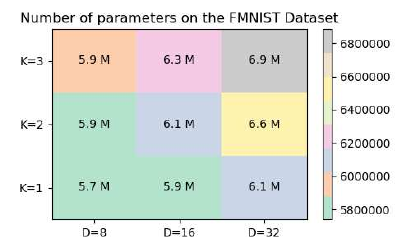}
  \caption{FMNIST number of parameters}
  \label{fig:sub2}
\end{subfigure}
\caption{CFC-CapsNet number of parameters for different choices of K and D}
\label{fig:nparam_param}
\end{figure*}
The impact of K and D on the number of parameters is reported in Figure \ref{fig:nparam_param}. For both datasets higher Ks result in a higher number of parameters. This is due to the enlargement of each FC layer as K grows. In addition as D increases, the number of parameters increases as well. The higher dimensionality the capsules have, the heavier the matrix multiplication process would be.

\par

\begin{figure*}
\centering
\begin{subfigure}{.5\linewidth}
  \centering
  \includegraphics[width=\linewidth]{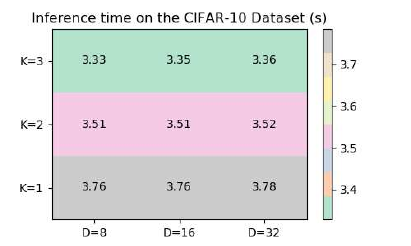}
  \caption{CIFAR-10 inference time}
  \label{fig:sub1}
\end{subfigure}%
\begin{subfigure}{.5\linewidth}
  \centering
  \includegraphics[width=\linewidth]{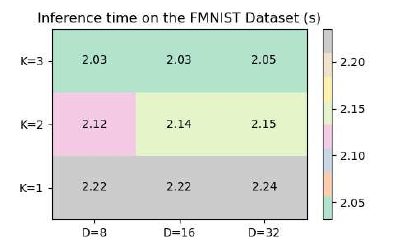}
  \caption{FMNIST inference time}
  \label{fig:sub2}
\end{subfigure}
\caption{CFC-CapsNet inference time for different choices of K and D}
\label{fig:test_param}
\end{figure*}

There is a slight change in the network inference time resulting from changing the values of D and K. Increasing K, leads to a network with fewer number of capsules. Consequently, it takes less time for the DR algorithm and the network becomes faster. However, increasing D slows down the network due to an increase in the complexity of the matrix multiplication process. 

\par
Considering the influences of the parameters K and D on the different aspects of the network's performance, we select $K=1$ and $D=8$ as the default values for the CFC layer. This choice achieves the minimum number of parameters while providing a competitive accuracy at the cost of a marginally lower speed. 

\subsubsection{CFC-CapsNet vs. CapsNet using the same budget}
In this section, we compare CFC-CapsNet with CapsNet using the same number of parameters. Even though the number of parameters in CFC-CapsNet can be further reduced by varying  K and D (as explained in the Parameter Exploration section), we continue to use $K=1$ and $D=8$. We reduce the number of parameters in CapsNet by reducing the number of kernels in the second convolutional layer. For CIFAR-10 dataset for 64 kernels, CapsNet uses 5.8M parameters that is close to the that of CFC-CapsNet (5.9M). Note that due to the method of implementation, the number of kernels should be divisible by 8\footnote{In this implementation, the output of the second convolutional layer (256 feature maps) is divided to 32 "capsule groups" each producing 8D capsules.}  . Therefore this is the closest we can get in terms of having the same number of parameters under the two architectures. 
\par
\begin{table}[htp]
\caption{CFC-CapsNet vs. CapsNet: Accuracy and the network time} 
\centering 
\resizebox{\linewidth}{!}{\begin{tabular}[]{|c|c| c| c|}
\hline\hline 
\textbf{Dataset} & \textbf{Training Time} & \textbf{Testing Time} & \textbf{Accuracy} \\ [0.5ex] 
\hline 

CapsNet & 50.98 & 7.78 & 68.21 \\ [1ex] 

\hline
CFC-CapsNet & 40.32 & 4.24 & 73.12 \\ [1ex]

\hline 
\end{tabular}}
\label{table:tres_samebug} 
\end{table}

As Table \ref{table:tres_samebug} shows, CFC-CapsNet gets 4.91\% higher accuracy and performs the training and testing faster than CapsNet.

\subsubsection{Network Speed-up}
\begin{figure}[htp]
    \centering
    \includegraphics[keepaspectratio]{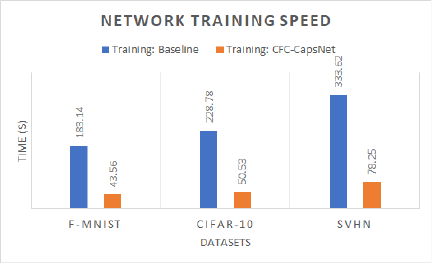}
    \caption{Network training time for CapsNet (left bar) and CFC-CapsNet (right bar). }
    \label{fig:res_train_speed}
\end{figure}

\begin{figure}[htp]
    \centering
    \includegraphics[keepaspectratio]{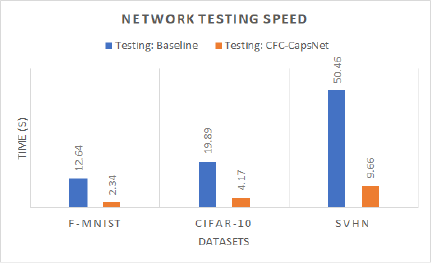}
    \caption{Network testing timefor CapsNet (left bar) and CFC-CapsNet (right bar). }
    \label{fig:res_test_speed}
\end{figure}

As a result of producing fewer number of capsules compared to CapsNet, CFC-CapsNet passes through the DR algorithm faster than CFC-CapsNet. Also, it includes fewer number of computations. The network training and testing times for both networks are shown in Figures \ref{fig:res_train_speed} and \ref{fig:res_test_speed} respectively. As shown in Figure \ref{fig:res_train_speed}, CFC-CapsNet is 4x faster than CapsNet for all datasets. Figure \ref{fig:res_test_speed} shows that inference is faster by 4.5 times in CFC-CapsNet compared to CapsNet for all datasets.

\newcommand{\specialcell}[2][c]{%
  \begin{tabular}[#1]{@{}c@{}}#2\end{tabular}}

\subsubsection{Network Accuracy and Number of Parameters}
\begin{table*}[htp]
\caption{CFC-CapsNet vs. CapsNet: Accuracy and the number of parameters} 
\centering 
\resizebox{\linewidth}{!}{\begin{tabular}[]{|c|c| c| c| c|c|}
\hline\hline 
\textbf{Dataset} & \specialcell{Parameters \\ (CapsNet)} & \specialcell{Parameters\\ (CFC-CapsNet)} & \specialcell{Accuracy \\ (CapsNet)} & \specialcell{Accuracy \\ (CFC-CapsNet)} & \specialcell{Accuracy \\  (CFC-CapsNet + Dropout)} \\ [0.5ex] 
\hline 

F-MNIST & 8.2M & 5.7M (-30\%) & 91.37\% & \textbf{92.86\%}  & 92.09\\ [1ex] 
\hline

CIFAR-10 & 11.7M & 5.9M (-49\%) & 71.69\% & 73.15\% & \textbf{73.85\%} \\ [1ex] 
\hline

SVHN & 11.7M & 5.9M (-49\%) & 92.70\% & \textbf{93.30\%} & 92.34\% \\ [1ex] 
\hline

\hline 
\end{tabular}}
\label{table:tres_acc} 
\end{table*}

We compare CFC-CapsNet (using K=1 and D=8) and CapsNet in terms of the network accuracy and the number of parameters in Table \ref{table:tres_acc}. As the table shows, CFC-CapsNet requires 30\% fewer weights for FMNIST and 49\% fewer parameters for the CIFAR-10 and SVHN datasets. In addition, CFC-CapsNet achieves a marginally higher accuracy compared to CapsNet for all the datasets. Regarding the capsule dropout, we used 0.4, 0.1 and 0.05 as the dropout rates. However, using dropout was beneficial only for the CIFAR-10 dataset. This can be explained as follows: since CFC-CapsNet includes only a few number of capsules, dropping capsules during training results in losing important information for the network.

\subsubsection{Comparison with state-of-the-art and DCNN}
In this section, we compare the classification accuracy of some of the state-of-the-art networks with that of CFC-CapsNet. It is important to note that CFC-CapsNet introduces the CFC layer as a universal method for summarizing the capsules and to compress the network, and it can be applied to any network based on CapsNet. In this work, CapsNet is taken as the baseline and the CFC layer is integrated into it, and it is shown that the integration of CFC layer is beneficial in all aspects of the network performance. We chose CapsNet (the first network based on capsules \cite{Sabour2017}) as a starting point to validate the CFC layer because it is a basic and robust network. The CFC layer requires modifications to be made before getting integrated into more complex networks based on CapsNet.

\par
Table \ref{table:tstofart} shows a list of state-of-the-art networks. Some recent DCNNs are shown on the top, and the recent CapsNet variants are shown in the middle. As the tables shows, networks based on CapsNet are still behind the DCNNs such as DenseNet \cite{Huang2016}, however they have achieved comparably high accuracy. At the bottom of the table, the results for CapsNet and CFC-CapsNet are shown. CFC-CapsNet introduces a layer to optimize any network based on CapsNet, and CFC-CapsNet is the result of customizing the CFC layer for CapsNet.

\begin{table}[htp]
\caption{The network test accuracy of state-of-the-art networks compared to CFC-CapsNet. Some recent DCNNs are shown on the top, and the recent CapsNet variants are shown in the middle. CFC-CapsNet introduces a layer to optimize any network based on CapsNet, and the results are corresponding to applying the layer to CapsNet.} 
\centering 
\resizebox{\linewidth}{!}{\begin{tabular}[]{|l|c| c|c|}
 \hline
\textbf{Model} & \textbf{CIFAR-10} & \textbf{SVHN} & \textbf{FMNIST} \\ [0.5ex] 
\hline 

DenseNet \cite{Huang2016}  & 96.40\% & 98.41\% & 95.40\%  \\ [1ex] 
RS-CNN \cite{Yang2020}  & 90.15\% & 95.56\% & 93.34\% \\ [1ex]
BiT-M \cite{Kolesnikov2019}  & 98.91\% & - & - \\ [1ex]
\hline
DA-CapsNet \cite{Huang2020} &  85.47\% & 94.82\% & 93.98\% \\ [1ex]
Cv-CapsNet++ \cite{He2019} & 86.70\% & - & 94.40\% \\ [1ex]
HitNet \cite{Deli2018} &  73.30\% & 94.50\% & 92.30\% \\ [1ex]
WR-CapsNet \cite{Fuchs2020}  & 93.43\% & 96.46\% & - \\ [1ex]
RS-CapsNet \cite{Yang2020}  & 91.32\% & 97.08\% & 94.08\% \\ [1ex]
DeepCaps (7-ens) \cite{Rajasegaran2019}  & 92.74\% & 97.56\% & 94.73\% \\ [1ex]
\hline
CapsNet (Pytorch) \cite{Sabour2017} & 71.69\% & 92.70\% & 91.37 \\ [1ex]
CFC-CapsNet  & 73.85\% & 93.30\% & 92.86\% \\ [1ex]

\hline 
\end{tabular}}
\label{table:tstofart} 
\end{table}

\subsubsection{Robustness to Affine Transformations}
To verify how robust CFC-CapsNet is against applying affine transformation to the input images, we follow the approach taken by Sabour et al. \cite{Sabour2017}. In this approach, the AffNIST dataset \footnote{https://www.cs.toronto.edu/~tijmen/affNIST/} is used. This dataset includes 32 different types of affine transformations applied to the MNIST dataset. The experiment involves training the network on the expanded MNIST dataset which includes only translations applied to the training and testing sets of the MNIST dataset \cite{LECUN}. Once the network achieves 99.23\% of accuracy, it is tested against transformed AffNIST dataset. Since this dataset includes 40x40 images, the expanded MNIST dataset is created by placing 28x28 images randomly on a 40x40 grid. There are 169 possible locations for each sample. 
\par
The Pytorch implementation we used for this work, achieves 90.52\% and 87.82\% on CapsNet and CFC-CapsNet respectively. The drop in the accuracy is because of using fewer capsules to represent the input image. Even though the few generated capsules are effective enough to provide a competitive representation of data on small-scale datasets, they fail to provide as accurate a viewpoint-invariant representation of the input image as that of CapsNet, and obtain a slightly lower accuracy. However, it is noteworthy that training and testing are 4.5x faster in CFC-CapsNet compared to that of CapsNet on the AffNIST dataset.

\section{Conclusion}
In this work, we proposed a new variant of CapsNet referred to as the CFC-CapsNet. This network enhances CapsNet by integrating a new layer called the CFC layer which is responsible for translating the low-level extracted features into capsules. This approach results in less parameters, faster training and inference and a slight increase in the network test accuracy. CFC-CapsNet achieves 50\% fewer parameters, four times faster inference and 2.16\% increase in the network accuracy for CIFAR-10 dataset. Tailoring the CFC layer further for other variants of CapsNet, and using consecutive CFC layers are some of the possible future paths for this research.


\begin{acknowledgements}
This research has been funded in part or completely by the Computing Hardware for Emerging Intelligent Sensory Applications (COHESA) project. COHESA is financed under the National Sciences and Engineering Research Council of Canada (NSERC) Strategic Networks grant number NETGP485577-15.
\end{acknowledgements}

%
%

\bibliographystyle{unsrt}{}   


%
%

\end{document}